\documentclass{article}

\PassOptionsToPackage{square, numbers}{natbib}
\usepackage[preprint]{neurips_2021}  

\usepackage[utf8]{inputenc} 
\usepackage[T1]{fontenc}    
\usepackage{hyperref}       
\usepackage{url}            
\usepackage{booktabs}       
\usepackage{amsmath, amssymb}
\usepackage{nicefrac}       
\usepackage{microtype}      
\usepackage{xcolor}         
\usepackage{subfigure}
\usepackage{graphicx}
\usepackage{algorithm, algorithmic}
\usepackage{bm}

\title{Exploiting the Hidden Tasks of GANs: \\ Making Implicit Subproblems Explicit}

\author{%
  Romann M.\ Weber \\
  DisneyResearch\textbar Studios\\
  Zurich, Switzerland \\
  \texttt{romann.weber@disneyresearch.com} \\
}

\begin{document}

\maketitle

\begin{abstract}
  We present an alternative perspective on the training of generative adversarial networks (GANs), showing that the training step for a GAN generator decomposes into two implicit subproblems.  In the first, the discriminator provides new target data to the generator in the form of \emph{inverse examples} produced by approximately inverting classifier labels.  In the second, these examples are  used as targets to update the generator via least-squares regression, regardless of the main loss specified to train the network.  We experimentally validate our main theoretical result and demonstrate significant improvements over standard GAN training made possible by making these subproblems explicit.  
\end{abstract}

\section{Introduction}
\label{intro}

Soon after their introduction, generative adversarial networks (GANs) \citep{goodfellow2014generative} quickly became the gold standard in implicit generative modeling.  In particular, when it comes to image generation, GANs generally achieve sharper and more convincing results than most of their non-adversarial counterparts (e.g.\ \citep{karras2020analyzing}).  Nevertheless, despite the flood of research into GANs and recent valuable insight into best practices for training these often temperamental models, a fundamental understanding of what makes GANs work so well under the hood remains elusive.

As suggested by their name, GANs are commonly thought to train \emph{adversarially}, pitting one network---a \emph{generator}---against another network---a \emph{discriminator}---in a tit-for-tat minimax game that ultimately settles into the saddle point of a Nash equilibrium \citep{goodfellow2016nips, lucic2017gans}.  A common personification of this game has the generator playing the role of a forger, who must learn to hoodwink an art expert with progressively better and better forgeries until the expert can no longer tell the difference.  Regardless of the exact tutorial example, the generator's assumed task of learning to ``fool'' the discriminator based on its feedback is widely accepted \citep{gui2020review}.

While it is clear that GANs receive training signal from their discriminators, the exact nature of discriminators' gentle nudging of generator output progressively closer to the target data distribution is both poorly understood and underexplored \citep{fedus2017many}. The portrayal of this process as a battle between discriminator and generator networks is both deliciously intuitive and pedagogically useful, but the perspective presented in this paper is that GAN training has hidden within it two simpler subproblems that may provide deeper insight into these models' inner workings.  We show that these subproblems are intimately connected to \emph{network inversion} and \emph{least-squares regression}.  

Specifically, we claim that GAN training is a fundamentally cooperative process in which the discriminator provides progressively better target data---in the form of \emph{inverse examples} created by approximately inverting classifier labels into the data domain---to the generator, which is then tasked with associating those targets with its inputs via squared loss, regardless of the main objective used to train the network as a whole.



The main contributions of this paper are the following:
\begin{itemize}
    \item We show that the training step of a GAN generator implicitly decomposes into two subproblems, a \emph{target-generation} problem and a \emph{regression} problem (\S \ref{sec: Theoretical}).
    \item We show that explicitly separating generator training into these subproblems introduces otherwise unavailable additional options into GANs, which can lead to significant improvements over standard training (\S \ref{sec: Experiments}).
\end{itemize}
In Appendix \ref{sec: Toy}, we also present an analysis of a toy example using our subproblem interpretation, which introduces a simple representation of the generator's \emph{inductive bias}, which we apply to describing the generator's output relative to its regression targets.

\section{Motivation} \label{sec: Motivation}

\begin{figure}[t]
\vskip 0.2in
\begin{center}
\includegraphics[height=2.5in]{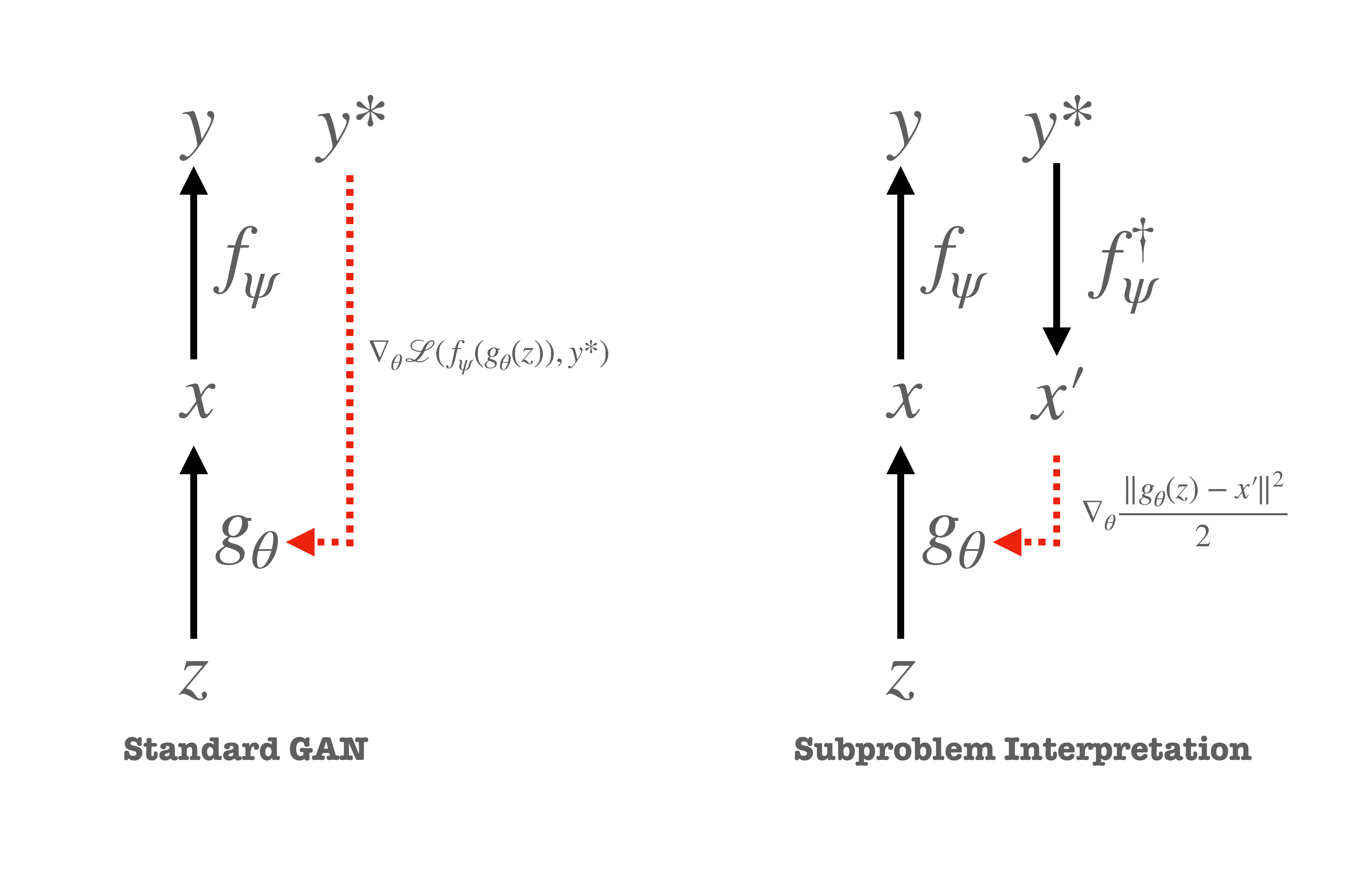}
\caption{A schematic representation of the standard viewpoint on GAN generator training versus the subproblem interpretation.  (See Algorithm \ref{alg:subproblem}.) An initial sample $x = g_{\theta}(z)$ is generated, which is likely classified by the discriminator as fake.  An \emph{inverse example} $x' = f_{\psi}^{\dagger}(y^*; x)$ (equation \eqref{eq: GenInverseUpdate}) is generated by approximately inverting the label $y^* = \text{``real''}$ through the discriminator into data space.  The generator is then updated via least-squares regression of input $z$ on the inverse example $x'$.}
\label{fig: schematic}
\end{center}
\vskip -0.2in
\end{figure}

The process by which we might associate an input $z \sim p(\mathcal{Z})$ with an output $x \sim p(\mathcal{X})$ lies at the heart of implicit generative modeling.  The distribution $p(\mathcal{Z})$ is often referred to as a \emph{prior distribution} and is typically something well modeled and easy to sample from.  Common choices for $p(\mathcal{Z})$ are the spherical Gaussian distribution and the uniform distribution in the unit hypercube.  The distribution $p(\mathcal{X})$ exists only indirectly as $p_{\text{data}}(\mathcal{X})$ in the form of the data on hand.  In most problems of interest, $p(\mathcal{X})$ follows no known functional form.

As $p(\mathcal{Z})$ and $p(\mathcal{X})$ have nothing to do with each other at the outset, there is no clear association to be made between a point randomly drawn from $p(\mathcal{Z})$ and any point in the training data set $\mathcal{X}$.  If there were, the modeling problem would reduce to a simple case of regression on paired data, and for any $z$ we could draw from $p(\mathcal{Z})$, we would have the ``perfect'' $x \in \mathcal{X}$ to serve as its target.  Instead, we use GANs to try to learn this association for us.

A standard GAN consists of the following two components: a generator, $g_{\theta}: \mathbb{R}^n \to \mathbb{R}^d$, and a discriminator, $f_{\psi}: \mathbb{R}^d \to \mathbb{R}$, which are parameterized by $\theta$ and $\psi$, respectively.  The goal in training a GAN is to take data $x \in \mathbb{R}^d \sim p(\mathcal{X})$ and train the generator to produce \emph{new} samples $x = g_{\theta}(z)$ that are plausibly drawn from $p(\mathcal{X})$ but \emph{not} part of the training data.

Previous analysis of GANs has suggested that they learn to approximate the data distribution by minimizing the Jensen-Shannon divergence between $p(\mathcal{X})$ and $p(g_{\theta}(\mathcal{Z}))$ during training \citep{goodfellow2014generative}.  However, this analysis relies on conditions that do not generally hold in practice \citep{DBLP:journals/corr/abs-1811-12402}.  Recent work has cast further doubt on this divergence-minimization assumption \citep{fedus2017many}, leaving the secret of GANs' success somewhat of a mystery.

When we break down the process of GAN training, particularly the training of the generator, we see that the training signal from the discriminator must ultimately serve to associate an input $z$ with a ``better'' output $x$ than it had produced prior to the generator update.  The standard approach to thinking about GANs says that this ``better'' $x$ comes from updating the generator parameters, $\theta$, to fool the discriminator.  

The alternative approach we propose says that the ``better'' $x$ actually comes \emph{first}, in the form of an implicit target for the generator to associate with $z$.  This target is produced by \emph{inverting a discriminator label into data space} and is then fit by the generator via \emph{least-squares regression}.  A visual schematic of the subproblem interpretation is shown in Figure \ref{fig: schematic}, and a pseudocode representation of our formulation is given in Algorithm \ref{alg:subproblem}.  We make the formal case for our viewpoint in Section \ref{sec: Theoretical}, following some additional background.

\section{Background} \label{sec: Background}

\subsection{Network Inversion} \label{sec: Inversions}

In the most general case, a neural network is a parameterized mapping $f: \mathcal{X} \subset \mathbb{R}^d \to \mathcal{Y} \subset \mathbb{R}^m$.  It is most commonly the case that $d \neq m$, meaning that the mapping is not bijective,\footnote{Even if $d = m$, there is no guarantee that the mapping is bijective.} so $f$ is not invertible in the traditional sense.

It is possible, however, to generalize our concept of what an inverse is if we allow the inverse mapping to simply be an association between sets as opposed to a well-defined function.  We define the \emph{generalized inverse} of $f$ as
\begin{equation}
    f^{\dagger}(y^*) = \{x \in \mathcal{X} | f(x) = y^* \}. \label{eq: GenInverse}
\end{equation}
In the case that $f$ is bijective, the definition \eqref{eq: GenInverse} corresponds to the traditional inverse.  Otherwise we have the set of all values in the domain that map to the target of interest.

The generalized inverse also lets us define a process for inverting functions that are otherwise not invertible.  Specifically, we can define a \emph{discrepancy} $\delta(y,y^*)$ to measure the disagreement between the target $y^*$ and candidate values $y = f(x)$.  We then identify the set of ``inverses'' of $y^*$ at all points $x$ such that $\delta(y=f(x), y^*) = 0$. We have considerable freedom in our choice of $\delta$, with clear options being $\ell_1$ and $\ell_2$, among others.  


If we cannot make the discrepancy completely vanish, we can still say to have found an \emph{approximate inverse} of $y^*$ in the domain of $f$ if the discrepancy is at least rendered tolerably small.  In any case, the problem to solve is 
\begin{equation}
    f^{\dagger}(y^*) = \arg \min_x \delta(f(x), y^*). \label{eq: Inversion}
\end{equation}
The approximate inversion of neural networks according to this formulation has received considerable attention in the literature (e.g.\ \citep{mahendran2015understanding, creswell2018inverting, xia2021gan}).

The iterative process by which one inverts a neural network's downstream representations or output into input space via \eqref{eq: Inversion} involves passing a gradient signal into a candidate input, $x$, which corresponds to an initial guess at the answer.  The process follows the familiar steps of gradient descent, one step of which is given by  
\begin{equation}
    x \leftarrow x - \lambda \nabla_x \delta(f(x),y^*), \label{eq: InverseUpdate}
\end{equation}
where $\nabla_x$ is the gradient with respect to $x$ and $\lambda$ is a small learning rate.

When the data domain $\mathcal{X}$ is highly structured, a na\"{i}ve application of \eqref{eq: InverseUpdate} can quickly provide a reminder of how troublesome many-to-one mappings can be.  With a purely random initial guess $x$, it is not unusual to arrive at a nearby ``inverse'' $x'$ that barely looks any different from $x$ but nevertheless almost exactly satisfies $f(x') = y^*$.  It is well known that in the case of classifiers, such imperceptible changes to input data can lead to completely different outputs.  When manufactured intentionally, these changes are known as \emph{adversarial attacks}, and these local spurious ``inverses'' are known as \emph{adversarial examples} \citep{goodfellow2014explaining}.  

In the network-inversion setting, it is common to introduce various regularization terms beyond the discrepancy $\delta$ in order to steer away from these undesired solutions and enforce the structure of the domain of interest.  For example, when dealing with natural images, \emph{total variation} (TV) is a common regularizer \citep{mahendran2015understanding}.  In fact, even the \emph{structure} of deep convolutional neural networks (CNNs) can function as a powerful regularizer when dealing with image data, even if the networks are randomly initialized and untrained \citep{ulyanov2018deep}.  This is due to such networks' \emph{inductive bias}, which refers to the set of assumptions implicit in a model or the class of functions the model can represent \citep{neyshabur2014search}.

Let $\delta_1$ represent the collection of any regularization terms and our discrepancy in a compound loss function.  The process of approximate inversion of $f(x) = y^*$ relative to the regularized discrepancy $\delta_1$ is then  
\begin{equation}
\begin{split}
    x' = f_{\psi}^{\dagger}(y^*;x) &=  x - \lambda_1 \nabla_x \delta_1 \\
    &= x - \lambda_1 D_x^\top (\delta_1), \label{eq: GenInverseUpdate}
\end{split}
\end{equation}
where
\begin{equation}
    D_x (\delta_1) = \left[ \frac{\partial \delta_1}{\partial x} \right] = \nabla_x^\top \delta_1 \label{eq: GenJacobian}
\end{equation}
represents a \emph{generalized Jacobian}\footnote{Without loss of generality, we can assume that all variables of interest have been vectorized, in which case $D$ represents the standard matrix Jacobian (which, it is important to note, we consider as the transpose of the gradient, $\nabla$).  However, the variables involved can be arbitrary tensors, which allows the generalized Jacobian to also be a tensor.  The key point is that the dimensions of $D$ are \emph{numerator (argument) by denominator (subscript)}.  So if $x \in \mathbb{R}^d$ and the loss $\mathcal{L}$ outputs a scalar, then $D_x(\mathcal{L}) \in \mathbb{R}^{1 \times d}$.  This generalizes easily to higher-order tensors $D$, with the concept of transposition stretched accordingly.} and $\top$ denotes matrix transposition.  Given the conceptual similarity to adversarial examples, going forward we will refer to the modified $x'$ from equation \eqref{eq: GenInverseUpdate} as an \emph{inverse example}.

\subsection{GAN Training} \label{sec: GANs}

GANs are famously tricky to train, and numerous methods have been proposed to stabilize the delicate training process (e.g.\ \citep{ganhacks, salimans2016improved}).  Among the first of these methods was a modification to the original cross-entropy formulation to avoid the loss from saturating and providing no useful gradient if the discriminator gets too good too fast \cite{goodfellow2016nips}, which is especially easy to have happen in the early stages of training.  Importantly, the results that follow do not depend on the exact nature of the loss used in training.

Let us examine a \emph{single step} of generator training using noise input drawn from $p(\mathcal{Z})$.  We will assume that our discriminator $f_{\psi}$ has completed a round of training and that its parameters, $\psi$, are frozen and untrainable.  Recall that during generator training, we are supplying the discriminator $f_{\psi}$ with labels saying that the samples produced by $g_{\theta}$ are real.  The goal of updating the generator parameters $\theta$ in this step is to adjust the output of $g_{\theta}$ to push the labels $y = f_{\psi}(g_{\theta}(z))$ closer to the target labels, $y^* = \text{``real''}$.  In other words, we want to minimize the discrepancy between the \emph{current} labels $y$ and the \emph{target} labels $y^*$.  That is, we have a loss defined by 
\begin{equation}
    \mathcal{L}_{g} = \delta (y=f_{\psi}(g_{\theta}(z)), y^*=\text{``real''}), \label{eq: GeneratorLoss}
\end{equation}
where the discrepancy $\delta$ can be binary cross-entropy \citep{goodfellow2014generative}, $\ell_2$ loss \citep{mao2017least}, or a number of other options. The update recipe to $\theta$ is straightforward, namely 
\begin{equation}
    \theta' = \theta - \eta_g D_{\theta}^\top(\mathcal{L}_{g}), \label{eq: ThetaUpdate}
\end{equation}
where $$D_{\theta}^\top(\mathcal{L}_{g}) = \left[ \frac{\partial \mathcal{L}_{g}}{\partial \theta} \right]^\top = \nabla_{\theta} \mathcal{L}_g$$ is the gradient of $\mathcal{L}_{g}$, expressed as the transposed generalized Jacobian.

Reducing the discrepancy between $y$ and $y^*$ does hint back at our earlier discussion of approximate inversion.  However, as the present goal is to update the generator \emph{parameters}, $\theta$, the connection is not necessarily obvious at this point.  We connect the dots in the following section.

\section{Theoretical Results} \label{sec: Theoretical}


Let us revisit the GAN generator loss \eqref{eq: GeneratorLoss}.  The process leading to this loss is given by the path 
\begin{equation}
    z \overset{g_{\theta}}{\longrightarrow} x \overset{f_{\psi}}{\longrightarrow} y \longrightarrow \mathcal{L}_g,
\end{equation}
with the Jacobian of the loss with respect to the generator parameters $\theta$ decomposing as the chain 
\begin{equation}
    D_{\theta}(\mathcal{L}_g) = D_{y}(\mathcal{L}_g)D_{x}(y) D_{\theta}(x). \label{eq: GenChain}
\end{equation}

Let us now assume that we have a new goal in mind, namely one of associating the input $z$ with the inverse example $x'$ as defined in equation \eqref{eq: GenInverseUpdate}.  We will define a simple squared loss as our discrepancy between this target and the generator's output $x(\theta) = g_{\theta}(z)$:
\begin{equation}
    \delta_2(x,x';\theta) = \frac{1}{2} \| x(\theta) - x' \|^2. \label{eq: SquaredLoss}
\end{equation}

The Jacobian of \eqref{eq: SquaredLoss} with respect to the parameters $\theta$ is given by 
\begin{equation}
    \begin{split}
        D_{\theta}(\delta_2) &= D_x(\delta_2) D_{\theta}(x) \\
        &= (x - x')^\top D_{\theta}(x) \\
        &= \lambda_1 D_y(\delta_1) D_x(y) D_{\theta}(x), \label{eq: ThetaUpdateFull}
    \end{split}
\end{equation}
where the final line follows directly from the second by rearranging and substituting \eqref{eq: GenInverseUpdate}, along with the decomposition $D_x(\delta_1) = D_y(\delta_1) D_x(y)$.  Equation \eqref{eq: ThetaUpdateFull} gives us what we need to update the generator parameters $\theta$ to push $x(\theta)$ toward $x'$, as its transpose defines the loss's gradient.  That gradient update is given by 
\begin{equation}
    \theta' = \theta - \lambda_2 D^\top_{\theta}(\delta_2). \label{eq: ThetaRegression}
\end{equation}

Combining \eqref{eq: ThetaRegression} and \eqref{eq: ThetaUpdateFull}, we have
\begin{equation}
\begin{split}
    \theta' &= \theta - \lambda_2 D^\top_{\theta}(x) (x - x') \\
        &= \theta - \lambda_1 \lambda_2 D^\top_{\theta}(x) D^\top_x(\delta_1) \\
        &= \theta - \lambda_1 \lambda_2 D^\top_{\theta}(\delta_1), \label{eq: identical}
\end{split}
\end{equation}
which is identical to the standard GAN generator update \eqref{eq: ThetaUpdate} when $\eta_g = \lambda_1 \lambda_2$ and $\mathcal{L}_g = \delta_1$.

The above results show that GAN generator training decomposes into two subproblems---generation of ``inverse examples'' followed by regression on these examples---that, for suitable choices of sub-objectives $\delta_1$ and $\delta_2$ and learning rates $\lambda_1$ and $\lambda_2$, will produce results identical to those of standard GAN training for \emph{any} GAN loss $\mathcal{L}_g$.  This suggests the following two questions:
\begin{enumerate}
    \item Does the subproblem interpretation admit flexibility into GAN training through alternative choices of $\delta_1$, $\lambda_1$,  $\delta_2$, and $\lambda_2$?
    \item Does this interpretation provide avenues for better understanding of GANs?
\end{enumerate}
The answer to the first question is unequivocally \emph{yes}, which the formulation presented so far and in Algorithm \ref{alg:subproblem} should make clear.  Less obvious is whether any of these alternative choices is \emph{better} than what is implicitly chosen by default in standard GAN training.  We address this question experimentally in Section \ref{sec: Experiments} and show that significant improvements over standard training are indeed possible.  We argue that the second question can be answered affirmatively as well and provide an example in this direction in Section \ref{sec: Toy}.

\section{Experimental Results} \label{sec: Experiments}

\begin{figure}%
\centering
\subfigure[Log-scale FID for various regression losses ($\ell_1, \ell_2$), inversion steps (1--5), and regression steps (R1, R2), reported in Section \ref{sec: factorial}.]{%
\label{fig:first}%
\includegraphics[height=2in]{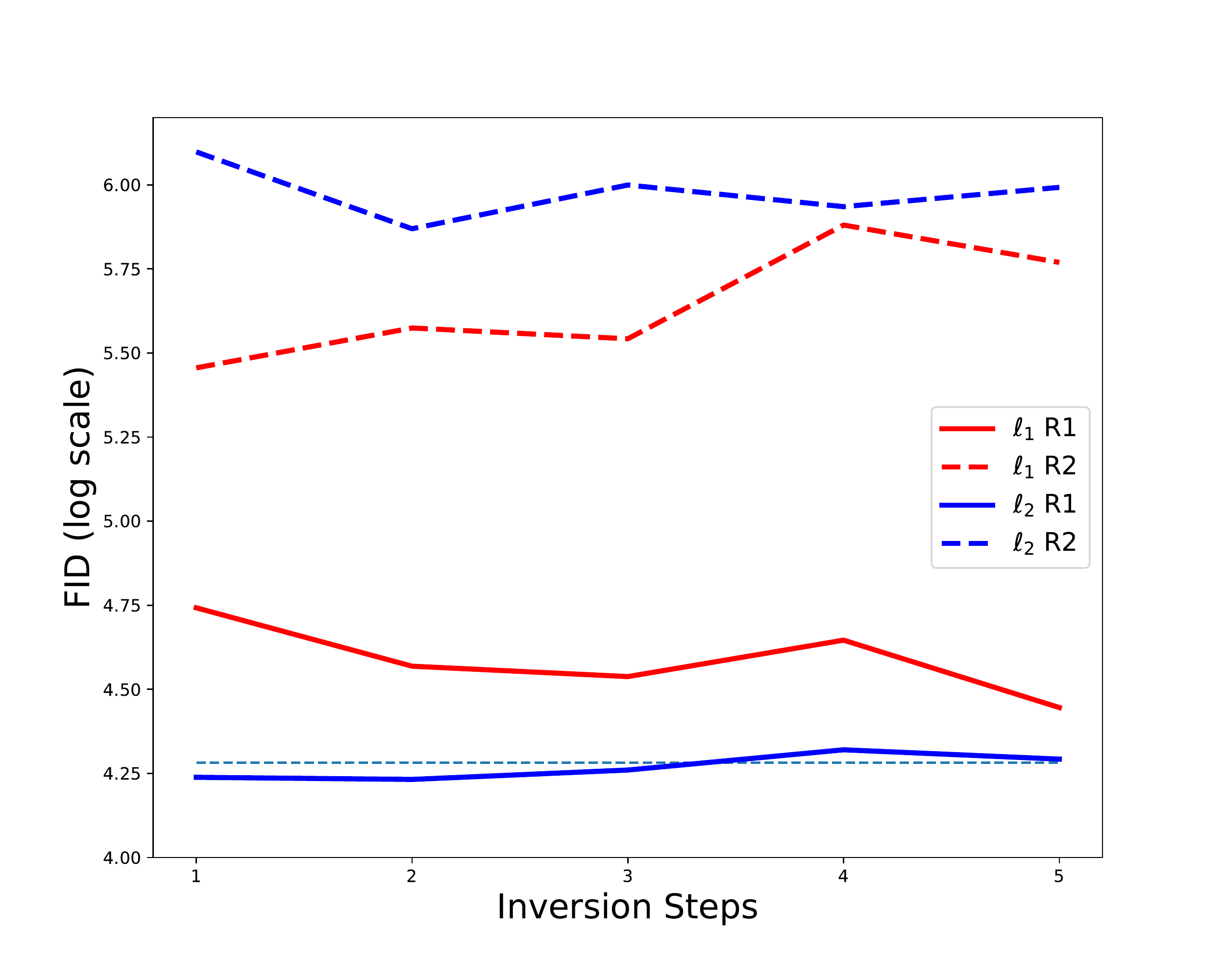}}%
\qquad
\subfigure[Average FID (with standard errors) versus inversion learning rate $\lambda_1$, reported in Section \ref{sec: rates}. Baseline indicated at top.]{%
\label{fig:second}%
\includegraphics[height=2in]{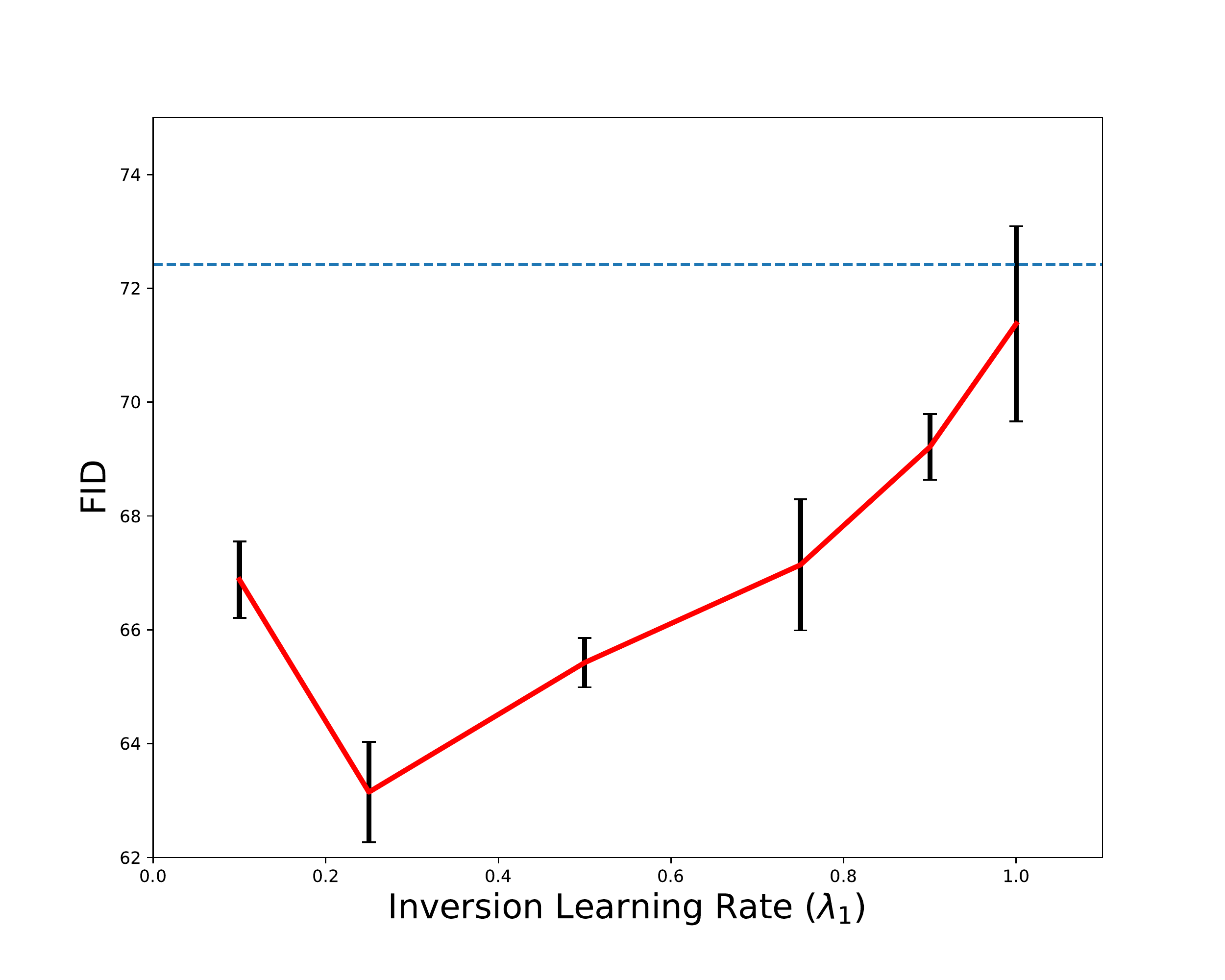}}%
\caption{The effect various training options on Fr\'{e}chet Inception Distance (FID).  In both figures, the dashed horizontal line indicates the default DCGAN baseline (FID = $72.42$) on the {\tt CelebA} data set.} \label{fig: experiments}
\end{figure}

A pseudocode representation of the subproblem formulation of GAN training is given in Algorithm \ref{alg:subproblem}.  The inputs to this algorithm show that among the options one has in this formulation that do not exist in standard GAN training are:
\begin{itemize}
    \item using an alternative discrepancy $\delta_1$ for inverting discriminator labels
    \item replacing the implicit $\ell_2$ regression loss with another loss (e.g.\ $\ell_1$, SSIM) in $\delta_2$
    \item introducing subproblem regularization as additional components of $\delta_1$ and $\delta_2$
    \item altering or otherwise scheduling the subproblem learning rates $\lambda_1$ and $\lambda_2$
    \item running additional inversion and/or regression steps during the generator update 
    \item using independent optimizers in each subproblem
\end{itemize}

A full demonstration of the possibilities introduced by the interpretation we propose is beyond the scope of this paper.  However, we present two sets of experiments to demonstrate our interpretation and some of its options.  We also compare our results to a standard implementation of DCGAN~\citep{radford2015unsupervised} on the {\tt CelebA} data set~\citep{liu2015faceattributes} with one modified to explicitly separate the inversion and regression subproblems.  Specifically, we used the official DCGAN implementation provided by PyTorch\footnote{\url{https://pytorch.org/tutorials/beginner/dcgan_faces_tutorial.html}} with all of its default parameters, along with the default random seed to initialize the networks.  

\begin{figure*}
\centering
\subfigure[Samples from the baseline DCGAN using default settings after 10 epochs (FID = $72.42$).]{\includegraphics[height=2.25in]{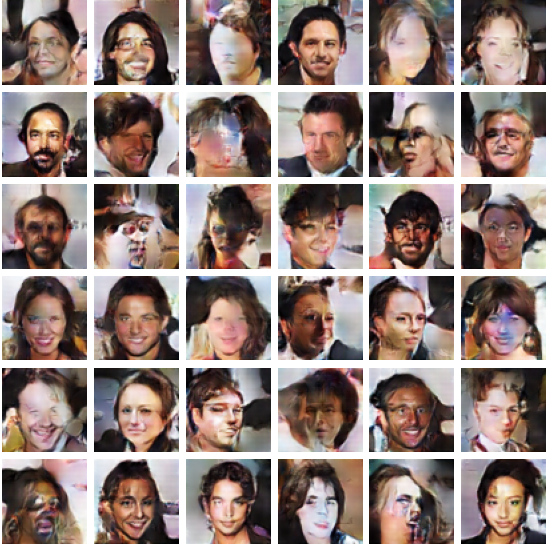}}  \qquad 
\subfigure[Samples after 10 epochs using the subproblem formulation with $\delta_2 = \ell_2$, $\lambda_1 = 0.25$, and $\lambda_2 = 0.0008$ (avg.\ FID = $63.15$).]{\includegraphics[height=2.25in]{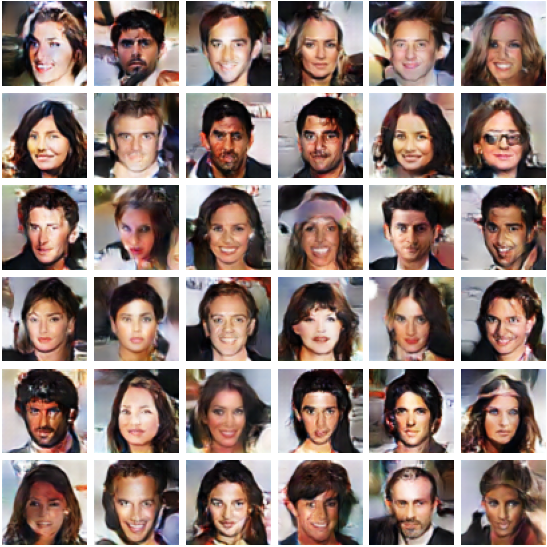}} 
\caption{Qualitative results comparing (a) standard GAN training and (b) the inversion-regression subproblem formulation introduced in this paper (see Algorithm \ref{alg:subproblem}).  Results were obtained in each case after roughly 40 minutes of training on the {\tt CelebA} data set. (Best viewed zoomed in.)}
\label{fig: training}
\end{figure*}

For our qualitative reference (Figure \ref{fig: training}(a)), we trained the PyTorch implementation of DCGAN with its default options on the {\tt CelebA} data set at $64 \times 64$ resolution.  We then edited the supplied code just enough to permit training in the subproblem regime, while keeping all other code and options set to their default values.  All training was performed on a single NVIDIA 1080Ti GPU.  

It is important to note that the focus of this demonstration is not on achieving state-of-the-art results in our generated samples.  Rather, the main points we wish to illustrate are that (1) the subproblem formulation derived above is indeed implicit in---and equivalent to---standard GAN training and (2) explicitly implementing the subproblem formulation admits training options that do not exist in the standard approach.

\subsection{Regression Losses, Inversion and Regression Steps} \label{sec: factorial}

We performed a three-way factorial experiment to examine the effect of regression loss ($\delta_2 \in \{ \ell_1, \ell_2 \}$), inversion steps ($N_1 \in [1,5]$), and regression steps ($N_2 \in [1,2]$) on Fr\'{e}chet Inception Distance (FID) \citep{heusel2017gans} relative to the DCGAN standard baseline.  In all cases, $\lambda_1 = 1/N_1$ and $\lambda_2 = 0.0002/N_2$.  (See Algorithm \ref{alg:subproblem} for a guide to notation.)  The standard DCGAN and the alternative versions were each trained from scratch from the same random seed for 10 epochs, for a total of 21 experiments.

After training, $2048$ images were generated from each GAN and compared with a random sample of $2048$ images from the {\tt CelebA} data set using a publicly available PyTorch implementation of FID \citep{Seitzer2020FID}.\footnote{The {\tt CelebA} data set, compared with itself using an identical procedure, has an FID of $13.70$.} Results for this set of experiments are shown in the left panel of Figure \ref{fig: experiments}.

The main takeaway from this set of experiments is that $\delta_2 = \ell_2$ as a regression loss with one regression step consistently outperformed other options, remaining near the DCGAN baseline FID of $72.42$.  Adjusting the regression learning rate to accommodate the scale of $\ell_1$ relative to $\ell_2$ may serve to close the performance gap between the losses, a matter we leave to future work.  In every case, however, adding additional regression steps ($N_2 > 1$) had a deleterious, occasionally catastrophic effect on output quality, while additional inversion steps left it largely unchanged or even slightly improved.

\subsection{Subproblem Learning Rates} \label{sec: rates}

Equation \eqref{eq: identical} shows that the subproblem formulation is equivalent to standard GAN training for certain choices of $\delta_1$, $\delta_2$, $\lambda_1$, and $\lambda_2$.  If one trains a GAN with learning rate $\eta_g$, then the subproblems should have rates such that $\eta_g = \lambda_1 \lambda_2$.  Beyond this relation, however, it is not obvious what the exact choices for these learning rates should be.

The default DCGAN implementation has a learning rate of $\eta_g = 0.0002$.  We tested combinations of $\lambda_1$ and $\lambda_2$ in our framework by choosing a set of values for the \textit{inversion learning rate}, $\lambda_1 \in \{ 0.1, 0.25, 0.5, 0.75, 0.9, 1 \}$ and correspondingly chose $\lambda_2 = \eta_g / \lambda_1$.  We performed one experiment in each condition with the default random seed and then repeated the experiments four times using fresh random seeds.  Average results for each condition, along with standard errors, are shown in the right panel of Figure~\ref{fig: experiments}.

In all cases, the tested options outperformed the baseline on average, showing significant improvements in all cases except $\lambda_1 = 1$, $\lambda_2 = \eta_g = 0.0002$, which we believe corresponds to the implicit default in the standard regime.  A qualitative comparison of samples from the best-performing setting is shown relative to the baseline in Figure \ref{fig: training}.



\section{Discussion and Concluding Remarks} \label{sec: Discussion}

The main result of Section \ref{sec: Theoretical} shows that hidden inside GAN training is an implicit cooperative process between a discriminator, which must learn progressively richer representations of the data, and a generator, which regresses noise inputs on data representations obtained by approximately inverting the discriminator.  By decomposing this process into explicit subproblems, we have shown that it is possible to improve over the standard approach by leveraging the additional control this alternative formulation affords.  To be sure, this additional control comes at a cost, namely in the form of additional choices to make during training.  Effectively, we have revealed a previously hidden panel of knobs and switches, and further research is needed to find the \emph{best} settings for them.


Since the apparent role of the discriminator, via implicit approximate inversion, is to nudge the generator distribution progressively nearer to the data distribution, it is worth asking whether this role can be filled by another method.  This is especially important if that method does not itself require training, as a major source of instability in GAN training is thought to arise from the nature of the minimax problem, or, more broadly, discriminators and generators whose individual objectives are at cross purposes \citep{gemp2018global, wang2019solving}.

Some recent work has attempted to tackle this problem. \citet{li2018implicit} perform \emph{implicit maximum likelihood estimation} to update generator parameters to minimize the sum of nearest-neighbor distances between generated samples and training data.  Follow-up work takes a similar approach but on distances in a reduced embedding space rather than in data space \citep{hoshen2019non}.  Importantly, the subproblem interpretation we present shows that the target-generation process can be separated from the regression process.  Without the need for a single process that is end-to-end differentiable, a variety of potential non-adversarial target-generation procedures become admissible.

The theoretical analysis we present assumes training via ``vanilla'' stochastic gradient descent, while it is more common in practice to train GANs using momentum-based optimizers that maintain a moving average of historical gradients.  Although this complicates the analysis somewhat, it does not change the main takeaway of the subproblem interpretation: At its heart, GAN training is about generating data-space targets to associate with noise inputs.  The presence of momentum in the optimization does not change this task, and we further note that the results we present in this paper were produced using momentum-based optimizers.

We conclude by suggesting that the additional control over generator training provided by the subproblem interpretation we propose makes a number of unresolved questions about GANs more approachable.  Our experimental results demonstrate that leveraging this additional control can lead to significant improvements relative to the default regime.  We see this paper as a first step toward a more thorough understanding of these extraordinarily powerful yet still-mysterious models.




\begin{algorithm}[tb]
  \caption{GAN Training Decomposed}
  \label{alg:subproblem}
\begin{algorithmic}
  \STATE {\bfseries Input:} data $x \sim p(\mathcal{X})$, noise $z \sim p(\mathcal{Z})$, discriminator $f_{\psi}$, generator $g_{\theta}$, discriminator loss $\mathcal{L}_f$, discriminator learning rate $\eta_f$, inversion discrepancy loss $\delta_1$, learning rate $\lambda_1$, inversion steps $N_1$ (default 1), regression loss $\delta_2$, learning rate $\lambda_2$, regression steps $N_2$ (default 1)
  \REPEAT
  \STATE Draw data $x \sim p(\mathcal{X})$ and noise $z \sim p(\mathcal{Z})$.
  \STATE Generate sample $\tilde{x} = g_{\theta}(z)$
  \STATE {\tt \# Update discriminator.}
  \STATE $\psi \leftarrow \psi - \eta_f \nabla_{\psi} \mathcal{L}_f( \{f_{\psi}(x), 1 \}, \{ f_{\psi}(\tilde{x}), 0 \} )$
  \STATE {\tt \# Create inverse examples.}
  \FOR{$i=1$ {\bfseries to} $N_1$}
  \STATE $\tilde{x} \leftarrow \tilde{x} - \frac{\lambda_1}{N_1} \nabla_{\tilde{x}} \delta_1 (\{ f_{\psi}(\tilde{x}), 1 \})$
  \ENDFOR
  \STATE {\tt \# Regress on inverse examples.}
  \FOR{$j=1$ {\bfseries to} $N_2$}
  \STATE $\theta \leftarrow \theta - \frac{\lambda_2}{N_2} \nabla_{\theta} \delta_2 (\{ g_{\theta}(z), \tilde{x} \})$
  \ENDFOR
  \UNTIL{Terminated}
\end{algorithmic}
\end{algorithm}

\bibliography{example_paper}
\bibliographystyle{plainnat}

\appendix


\section{A Toy Example} \label{sec: Toy}

Let us consider one of the simplest GAN setups we can imagine, namely one in which our generator is a one-layer linear neural network (equivalent to an affine transformation) and our discriminator is also a one-layer neural network with a sigmoid output (equivalent to logistic regression).  Let our data distribution be $p(\mathcal{X}) = \mathcal{N}(\bm{\mu}, \bm{\Sigma})$ and our prior distribution be $p(\mathcal{Z}) = \mathcal{N}(0,\mathbf{I})$, with both $\mathcal{Z}$ and $\mathcal{X}$ in $\mathbb{R}^d$.  In this case, we know that in order to approximate the data distribution, our generator needs to learn the mapping $g(z) = \bm{\Sigma}^{\frac{1}{2}}z + \bm{\mu}$, where $\bm{\Sigma}^{\frac{1}{2}}$ denotes the matrix square root.

For our discriminator, we have $f(x) = \sigma(\mathbf{w}^\top x)$, where $$\sigma(s) = \frac{1}{1 + \exp (-s)}$$ is the logistic function and $\mathbf{w} \in \mathbb{R}^d$.  For our generator, we have $x = \mathbf{B}\tilde{z}$, where $\tilde{z}$ is our input vector augmented with an extra dimension holding the constant value 1 and $\mathbf{B} \in \mathbb{R}^{d \times (d+1)}$.  Let us break down the training of this simple GAN.

\subsection{Discriminator Training}

We will assume that we are using binary cross-entropy as our discriminator loss function.  Let $v = \mathbf{w}^\top x$.  Our loss for an example $x$ is then given by 
\begin{equation}
    \mathcal{L}_{f}(x) = -y^* \log (\sigma(v)) - (1-y^*) \log (1 - \sigma(v)), \label{eq: BCE}
\end{equation}
where $y^*$ represents the ground-truth label for an example.  The gradient of this function with respect to $\mathbf{w}$, the parameter we need to optimize, is 
\begin{equation}
    \nabla_{\mathbf{w}} \mathcal{L}_{f} = (\sigma(v) - y^*) x. \label{eq: DiscGradw}
\end{equation}

With labels $y^* \in \{0, 1 \}$, we have $(\sigma(v) - y^*) \geq 0$ when $y^* = 0$ and $(\sigma(v) - y^*) \leq 0$ when $y^* = 1$.  Since we are optimizing through gradient descent, it winds up being the case that in expectation  
\begin{equation}
    \mathbf{w} \leftarrow \alpha \bar{X_1} - \beta \bar{X_0},
\end{equation}
where $\alpha, \beta \geq 0$ and $\bar{X_1}, \bar{X_0}$ represent, respectively, the empirical averages of the positive (real) and negative (generated) class examples.  In short, we tend toward learning a weight vector roughly equal to a weighted difference of class means.\footnote{That we would learn such a weight vector makes sense.  Consider restricting $\mathbf{w}$ to $\mathbb{S}^{d-1}$, the unit hypersphere in $\mathbb{R}^d$.  Then $$\arg \max_{\mathbf{w \in \mathbb{S}^{d-1}}} \mathbb{E} \left[ \mathbf{w}^\top (X_1 - X_0) \right] = \frac{\bar{X_1} - \bar{X_0}}{\|\bar{X_1} - \bar{X_0}\|}.$$}

\subsection{Generator Training}

We now try to optimize $\mathbf{B}$ by supplying bogus labels $y^* = 1$ for our generated samples.  As we showed earlier, this process is equivalent to first defining a set of ``inverse examples'' in the neighborhood of the current samples that more closely satisfy $f(x) = y^*$.  We create these examples by performing gradient descent on \eqref{eq: BCE} with respect to $x$, now with $y^* = 1$. 

The gradient of \eqref{eq: BCE} with respect to $x$ is given by 
\begin{equation}
    \nabla_{x} \delta_1 = \nabla_{x} \mathcal{L}_{f} = (\sigma(v) - y^*) \mathbf{w}. \label{eq: DiscGradx}
\end{equation}
Since $(\sigma(v) - y^*) \leq 0$ in this case, gradient descent on an example $x$ leads to 
\begin{equation}
    x' = x + \lambda_1 k \left( \alpha \bar{X_1} - \beta \bar{X_0} \right) \label{eq: MakeInverseExample}
\end{equation}
for $k > 0$.  That is, we are nudging $x$ away from the mean of the generated data and toward the mean of the real data.

\subsection{Analysis} \label{sec: analysis}

Returning to the general case of standard GAN training, we might ask what the generator would produce \emph{after} the parameter update \eqref{eq: ThetaUpdate} in response to the same input, $z$.  Since we assume that the learning rate $\eta_g$ is small, the change in the output can be well approximated by a first-order Taylor polynomial as
\begin{equation}
\begin{split}
    g_{\theta'}(z) &\approx g_{\theta}(z) + \nabla_{\theta}^\top g_{\theta}(z) \Delta \theta \\
    &= x + \left[ D_{\theta} (x) \right] \left[ -\eta_g D_{\theta}^\top (\mathcal{L}_g) \right] \\
    &= x - \eta_g D_{\theta} (x) D_{\theta}^\top (\mathcal{L}_g). \label{eq: NextGanOutput}
\end{split}
\end{equation}

If we substitute in the decomposition \eqref{eq: GenChain} into the third line of \eqref{eq: NextGanOutput}, we obtain 
\begin{equation}
    \begin{split}
        x'_{\text{GAN}} &= x - \eta_g D_{\theta}(x) D_{\theta}^\top(x) D_x^\top(y) D_y^\top(\mathcal{L}_g)  \\
        &= x - \eta_g \left[ K_{\theta}(x) \right] D_x^\top (\mathcal{L}_g), \label{eq: NextGanOutputFull}
    \end{split}
\end{equation}
since $D_x^\top (\mathcal{L}_g)  = D_x^\top(y) D_y^\top(\mathcal{L}_g)$, and where 
\begin{equation}
    \left[ K_{\theta}(x) \right] = D_{\theta}(x) D_{\theta}^\top(x) =  \left[ \frac{\partial x}{\partial \theta} \right] \left[ \frac{\partial x}{\partial \theta} \right]^\top \label{eq: Kernel}
\end{equation} is a $d \times d$ structure that describes the pairwise similarity of the features of $x$ as measured by the inner products of their partial derivatives with respect to the model parameters.


An interpretation of the role of $K_{\theta}(x)$ comes from viewing the last line of \eqref{eq: NextGanOutputFull} as its own first-order Taylor polynomial.  Since $\Delta x = - \eta_g D_x^\top(\mathcal{L}_g)$, we can see that 
\begin{equation}
    \left[ K_{\theta}(x) \right] \approx \left[ \frac{\partial g_{\theta}(z)}{\partial x} \right],
\end{equation}
and since $g_{\theta}(z) = x$, it is clear that $K_{\theta}$ should tend to the identity if $g_{\theta}$ has sufficient capacity.  In this way, $K_{\theta}(x)$ encodes the \emph{inductive bias} of the generator $g_{\theta}$, as it transforms the ``request'' made by $D_x^\top(\mathcal{L}_g)$ to update $x$ based on the network's ability to represent it.

In our toy example, since $x = g_{\theta}(z) = \mathbf{B} \tilde{z}$, we have 
\begin{equation}
    \mathbf{J} = \left[ \frac{\partial x}{\partial \theta} \right] = \left[ \frac{\partial \mathbf{B} \tilde{z}}{\partial \mathbf{B}} \right] \in \mathbb{R}^{d \times d \times (d + 1)},
\end{equation}
a rank-three tensor.  The entries of this tensor are given by 
\begin{equation}
    \mathbf{J}_{ijk} = \begin{cases} \tilde{z}_k & i = j \\ 0 & i \neq j \end{cases}.
\end{equation}
We can then calculate the $d \times d$ structure \eqref{eq: Kernel} as
\begin{equation}
    K_{\mathbf{B}}(x) = \mathbf{JJ}^\top = \mathbf{J} \ast_{(ilm, jlm, ij)} \mathbf{J},
\end{equation}
where the rightmost expression employs a generic tensor multiplication operator \citep{laue2020simple}.  Completing the calculation yields a diagonal matrix $K_{\mathbf{B}}(x)$ whose nonzero entries are $\| \tilde{z} \|^2$, with an expected value of $d + 1$ for a standard normal $z$.\footnote{Recall that we have augmented $z$ with a constant to form $\tilde{z}$.}

Since $K_{\mathbf{B}}(x)$ is proportional to the identity, the generator is capable of representing any target the discriminator provides as an inverse example.  However, inspection of \eqref{eq: MakeInverseExample} should make it clear that this GAN design has no hope of matching the target covariance $\bm{\Sigma}$ and can only align the \emph{means} of the generated and target data, which is straightforward to confirm experimentally.  This speaks to the larger question of the capacity of various discriminators to distinguish between distributions with important higher moments.\footnote{Also key are the underlying data representations, which can in principle be sufficiently rich to encode higher moments.  Consider, for instance, the mapping $x \to \exp (i \langle \mathbf{t}, x \rangle )$ for different choices of $\mathbf{t}$.  In this case, the resulting class means would correspond to the respective distributions' characteristic functions evaluated at $\mathbf{t}$.}

\end{document}